\documentclass[lettersize,journal]{IEEEtran}
\usepackage{amsmath,amsfonts}
\usepackage{algorithmic}
\usepackage{algorithm}
\usepackage{flushend}
\usepackage{array}
\usepackage{fontawesome5}
\usepackage[colorlinks=true, linkcolor=red, citecolor=ngreen, urlcolor=blue]{hyperref}
\usepackage[caption=false,font=normalsize,labelfont=sf,textfont=sf]{subfig}
\usepackage{textcomp}
\usepackage{multirow} 
\usepackage{stfloats}
\usepackage{url}
\usepackage{verbatim}
\usepackage{graphicx}
\usepackage{relsize}
\usepackage{ragged2e}
\usepackage{pgfplots}
\usepackage{colortbl}
\usepackage{cite}
\usepackage{float}
\usepackage{dsfont}
\usepackage{arydshln}
\usepackage{enumitem}
\usepackage{float}
\usepackage{hyperref}
\usepackage{multirow}
\usepackage{caption}
\usepackage{amssymb}
\usepackage{booktabs}
\usepackage{bbm}
\usepackage{tikz}
\usepackage{bm}
\usepackage[normalem]{ulem}
\usepackage{xcolor}
\usepackage{orcidlink} 
\usepackage{hyperref}
\hypersetup{
    pdfborder={0 0 0} 
}
\definecolor{mycolor_green}{RGB}{86, 170, 86}
\definecolor{mycolor_red}{RGB}{255, 103, 103}
\definecolor{mycolor_yellow}{RGB}{190, 190, 0}
\definecolor{fmycolor_green}{RGB}{88, 142, 49}
\definecolor{fmycolor_red}{RGB}{162, 24, 40}
\definecolor{fmycolor_blue}{RGB}{72, 116, 203}

\pgfplotsset{compat=1.18} 
\begin{document}
\definecolor{ngreen}{RGB}{17, 173, 30}
\title{\texttt{CSTA}: Spatial-Temporal Causal Adaptive Learning for Exemplar-Free Video Class-Incremental Learning}

\author{Tieyuan Chen$^{\orcidlink{0009-0005-7939-7139}}$, Huabin Liu$^{\orcidlink{0000-0001-9174-1696}}$, Chern Hong Lim$^{\orcidlink{0000-0003-4754-6724}}$,~\IEEEmembership{Senior Member, IEEE}, John See$^{\orcidlink{0000-0003-3005-4109}}$,~\IEEEmembership{Senior Member, IEEE}, \\Xing Gao$^{\orcidlink{0000-0001-8458-7483}}$, Junhui Hou$^{\orcidlink{0000-0003-3431-2021}}$,~\IEEEmembership{Senior Member, IEEE}, Weiyao Lin$^{\orcidlink{0000-0001-8307-7107}}$,~\IEEEmembership{Senior Member, IEEE}
\thanks{Tieyuan Chen, Huabin Liu, and Weiyao Lin are with the Department of Electronic Engineering, Shanghai Jiao Tong University, Shanghai, China. Tieyuan Chen and Weiyao Lin are also with the Zhongguancun Academy, Beijing, China. (e-mail:
\{tieyuanchen, huabinliu, wylin\}@sjtu.edu.cn).}
\thanks{Chern Hong Lim is with the School of Information Technology, Monash University, Selangor, Malaysia. (e-mail: lim.chernhong@monash.edu).}
\thanks{John See is with the School of Mathematical and Computer Sciences, Heriot-Watt University, Malaysia Campus, Malaysia (email: j.see@hw.ac.uk)}
\thanks{Xing Gao is with the Shanghai Artificial Intelligence Laboratory, Shanghai, China. (email: gxyssy@163.com).}
\thanks{Junhui Hou is with the Department of Computer Science, City University of Hong Kong, Kowloon Tong, Hong Kong. (email: jh.hou@cityu.edu.hk).}}

\markboth{IEEE TCSVT Submission}%
{Chen \MakeLowercase{\textit{et al.}}: CSTA}


\maketitle

\begin{abstract}
Continual learning aims to acquire new knowledge while retaining past information. Class-incremental learning (CIL) presents a challenging scenario where classes are introduced sequentially. 
For video data, the task becomes more complex than image data because it requires learning and preserving both spatial appearance and temporal action involvement. 
To address this challenge, we propose a novel exemplar-free framework that equips separate spatiotemporal adapters to learn new class patterns, accommodating the incremental information representation requirements unique to each class.
While separate adapters are proven to mitigate forgetting and fit unique requirements, naively applying them hinders the intrinsic connection between spatial and temporal information increments, affecting the efficiency of representing newly learned class information. 
Motivated by this, we introduce two key innovations from a causal perspective. 
First, a causal distillation module is devised to maintain the relation between spatial-temporal knowledge for a more efficient representation. 
Second, a causal compensation mechanism is proposed to reduce the conflicts during increment and memorization between different types of information. 
Extensive experiments conducted on benchmark datasets demonstrate that our framework can achieve new state-of-the-art results, surpassing current example-based methods by \textbf{4.2\%} in accuracy on average.
\end{abstract}

\begin{IEEEkeywords}
Action recognition, Class-incremental learning, Causal inference.
\end{IEEEkeywords}

\section{Introduction}
\label{sec:intro}
Class-incremental learning (CIL) presents a challenging continual learning scenario, where models are sequentially trained on new classes over data increments. The model needs to learn representations for the novel classes at each new increment while avoiding catastrophic forgetting of old classes learned in previous increments. 
In video CIL, due to the peculiarity of incorporating multi-frame information, the difficulties not only involve maintaining spatial knowledge of object appearances but also preserving temporal knowledge concerning how actions evolve~\cite{csvt2, csvt3, csvt1, he}. 

Most popular video CIL methods adopt an exemplar-based paradigm, storing a subset of prior task samples in memory for replay during new task learning. While effective in mitigating forgetting, replaying entire videos is impractical due to high storage costs compared to images. Approaches like selecting individual frames or applying compression~\cite{framemaker,vclimb,glimpse2023} reduce storage demands but often neglect temporal dynamics, leading to a loss of temporal information.
Beyond these, exemplar-free methods, such as CLIP-based approaches~\cite{st_prompt}, use pre-defined prompt pools but depend heavily on large-scale pre-training. Recently, STSP~\cite{stsp} proposed replacing linear classifiers with subspace classifiers, reducing exemplar reliance by storing representative features. However, the dominance of linear classifiers in classification models~\cite{csvt8,csvt9,csvt10} limits the generalization of this method.

\begin{figure*}
  \centering
   \includegraphics[width=0.95\linewidth]{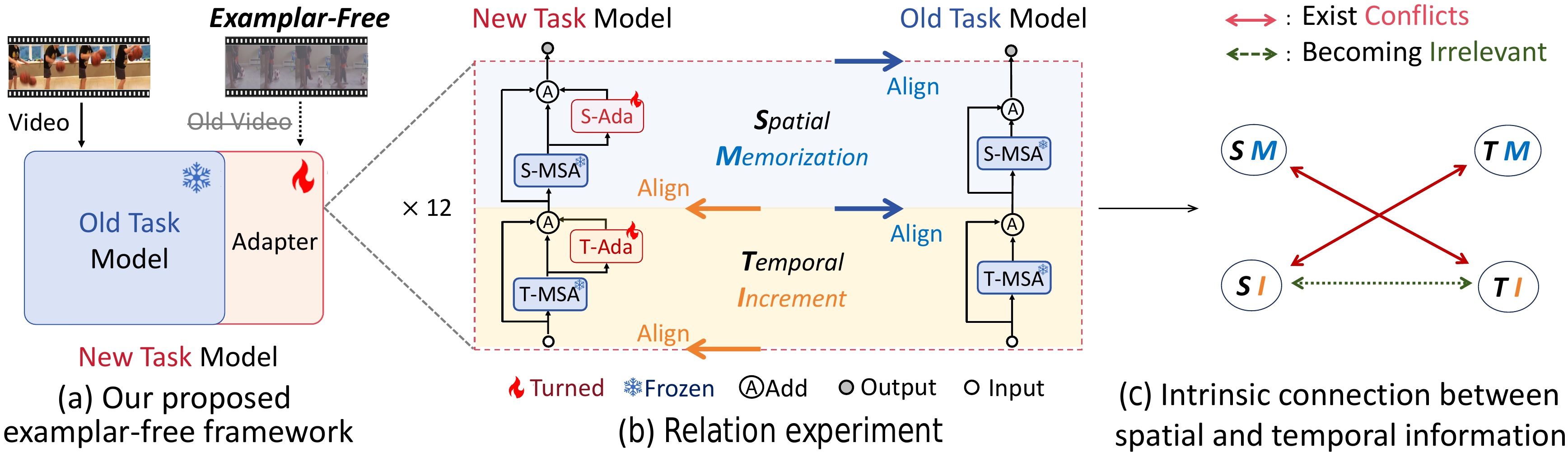}
  \caption{\textbf{Framework and Relation Experiment.} In sub-figure (a), our method utilizes lightweight adapters to accomplish an examplar-free framework. In sub-figure (b), to reach a more efficient representation during adaptation, we analyze the relation between spatial/temporal increment and memorization by making a comparison between introducing an adapter or not, where S-MSA and T-MSA represent spatial and temporal multi-head self-attention, S-Ada and T-Ada represent adaptation modules, alignment indicates aligning classification results with and without adaptation via KL-loss. Through analysis of the optimization directions, as shown in sub-figure (c), we observe that spatial and temporal increments become increasingly irrelevant, and conflicts emerge between the increment of one’s knowledge and the memorization of the other’s.}
  \label{fig:intro}
\end{figure*}

In summary, existing exemplar-based methods require substantial storage space and fail to retain the knowledge of learned action evolution. Conversely, exemplar-free approaches, such as those based on models like CLIP, depend heavily on large-scale pre-training and often exhibit limited generalization capabilities.
To tackle these challenges, we introduce an exemplar-free framework for video class-incremental learning, which does not rely on extensive pre-training and demonstrates robust generalization capabilities.
Our framework utilizes dynamic network architectures, introducing dedicated adaptable parameters to represent new knowledge while freezing the parameters storing previously learned knowledge. 
Specifically, at each increment, we newly introduce adapters with trainable parameters to each block of the base video transformer model (e.g. TimeSformer~\cite{timesformer}). 

As discussed above, the new knowledge required to be represented consists of temporal and spatial information, with different necessities for representation increment depending on the context.
The requirements actually depend on the relationship between previously acquired action categories and those newly learned.
For instance, actions such as kicking a soccer ball and beach soccer exhibit similar temporal action evolution (kicking actions) despite differing spatial visual features (playing fields). 
Conversely, actions like moving objects from left to right versus right to left involve opposite temporal dynamics yet share similar spatial features. 
Therefore, we design separate adapters for temporal and spatial information to address varied requirements, as depicted in Fig.~\ref{fig:intro} (a).

Specifically, at each increment, we newly append a temporal adapter and a spatial adapter with trainable parameters. 
These adapters learn spatial and temporal patterns associated with the new classes respectively, achieving class increments without disturbing the existing knowledge within the base model. 

However, while meeting the unique needs of spatial and temporal increment, it is natural to wonder whether the separation of spatiotemporal attention and adapters might gradually lead to a weakening of their relationship, making it difficult to maintain a unified and comprehensive representation of action categories.
Therefore, we explore the distinct features of spatial and temporal increments~(detailed in the middle of Fig.~\ref{fig:intro} and Sec.~\ref{Sec:analyze}), and note that adopting this spatiotemporal separated dynamic network approach presents the following two primary challenges:
\begin{itemize}
    \item[(1)] Separated adaptation will impair the inherent relation between spatial and temporal knowledge.
    \item[(2)] Conflicts arise between the increment of one's knowledge and the memorization of the other's.
\end{itemize}


Inspired by the intrinsic causality between spatial and temporal patterns, we further introduce two novel components from a causal perspective to our framework to address the above issues. 
Specifically, we first devise a causal distillation loss to maintain the coupling between the spatial and temporal adapters by distilling causal correlations on new data. This strengthens their spatial-temporal knowledge relation. 
Moreover, a causal compensation mechanism is proposed to reduce the conflicts between spatial increment (or memorization) and temporal memorization (or increment). This is accomplished by enhancing the benefits within the progress of the other information memorization.

We validate our approach through extensive experiments on benchmark video datasets, demonstrating that our proposed dynamic example-free framework can outperform previous SOTA by up to \textbf{4.2\%} on average while saving \textbf{61.9\%} storage budget. Overall, our key contributions are:

\begin{itemize}
    \item We present the first \emph{exemplar-free} robust framework without large-scale pre-training for video CIL.
    \item We propose a \emph{lightweight} adapter-based dynamic network that achieves both knowledge maintenance and increment by only introducing manageable parameters.
    \item We reveal the relations between temporal and spatial information during increment and propose causal methods to maximize benefits while minimizing conflicts.
\end{itemize}

\section{Related work}
\label{sec:2_related_work}
\subsection{Class-Incremental Learning}
Over the past few years, numerous studies have systematically explored CIL from three primary perspectives: data replay, knowledge distillation, and dynamic network expansion. 
Data replay methods, as exemplified by~\cite{data1, data2, data3}, focus on addressing CIL through the use of exemplars.  
This approach is straightforward for combating forgetting, but it requires privacy permissions and significant storage space. 
Consequently, subsequent research has increasingly sought alternative methods. 
One of the mainstream methods, knowledge distillation methods~\cite{kd1, kd2, kd3} utilizes knowledge distillation to resist forgetting or rectify the bias in the CIL model. 
Distillation-based methods combine the regularization loss (e.g. KL-loss) with the standard classification loss to update the network. The regularization term is calculated between the old and new networks to preserve previous knowledge when learning new data.
However, distillation may not always be aligned with the learning of new classes and can sometimes even hinder performance~\cite{hu2021distilling}.
The other mainstream method, dynamic network methods, as reported in~\cite{dy1, dy2, dy3}, involves expanding the network structure to enhance its representation capabilities. 
Dynamic network-based models assume the network’s capacity is finite and propose expanding the network when needed to enhance representation ability. 
However, the respective disadvantage of the dynamic network is the continuous increase in model parameters.

\subsection{Video Class-Incremental Learning}
Video CIL has been a popular research topic in the past few years. 
In the earliest batch of research works, TCD~\cite{tcd}, HCE~\cite{hce}, and Teacher-Agent~\cite{teacher} have proposed an idea of storing video and utilizing feature distillation and other similar supervision. 
Based on these studies, Scholars have begun to explore how to perform more efficient data replay. FrameMaker~\cite{framemaker} has proposed to store a single frame compression exemplar of video for data replay. 
vCLIMB~\cite{vclimb} has provided the first standard test-bed for video CIL and proposes to store down-sampling exemplars. 
And SMILE~\cite{glimpse2023} proposed to only store a simple frame of video. 
However, all methods above have privacy issues and temporal information is lost to a large extent. 
Furthermore, ST-prompt~\cite{st_prompt} achieves examplar-free by employing the CLIP model to match corresponding prompts from predefined prompt pools, however, it relies on large-scale pre-training and requires predefined pools as well. 
STSP~\cite{stsp} proposed a discriminative temporal-based subspace classifier that represents each class with an orthogonal subspace basis and adopts subspace projection loss for classification. 
However, STSP has replaced the linear classifier with a new subspace classifier, which may affect its generalization when applied to other models.
Therefore, we are committed to proposing a robust examplar-free method that effectively preserves temporal information without requiring large-scale pre-training.


\subsection{Dynamic Networks in CIL}
Given the finite capacity of a trained model, naively adapting to new knowledge will result in forgetting previous knowledge~\cite{yan2021dynamically}. 
To this end, the dynamic network is introduced to improve the model's representation ability dynamically by introducing new parameters. 
In the image CIL,  DER~\cite{yan2021dynamically} expands a new backbone when encountering new knowledge and finally aggregates the output of backbones. 
FOSTER~\cite{foster} further introduces knowledge distillation to save the number of newly introduced parameters. 
MEMO~\cite{memo} reveals that shallow layers of models trained on different knowledge share similarities, while deep layers exhibit diversity. Consequently, MEMO only expands deep layers. 
Despite the effectiveness of the dynamic network, its primary drawback is the significant increase in the number of model parameters. 
Therefore, we aim to integrate the adapter fine-tuning technique with dynamic network methods to mitigate this issue.


\subsection{Causal Inference}
Causal inference seeks to establish reasonable causal relations between factors, enabling the analysis of the effects of different treatments on specific factors. This approach is widely utilized in vision-related fields, e.g. visual question answering~\cite{niu2021counterfactual}, image segmentation~\cite{zhang2020causal}, video reasoning~\cite{mecd}. As for class incremental learning, the causal inference-based method mainly includes the following two works. 
DDE~\cite{hu2021distilling} introduces an anti-forgetting causal distillation method to distill the causal effect of previous knowledge, aiming to approximate data reply.
CafeBoost~\cite{qiu2023cafeboost} acknowledges the task order as a task-induced bias and causal inference is utilized to cut off the corresponding causal relation. 
However, the above causal inference methods analyze previous knowledge and current knowledge in coarse terms. 
In our work, we reveal the intrinsic causal relations between previous and current knowledge by spatial-temporal decomposition. 
Moreover, unlike previous methods that introduce causal inference through distillation, ours is through a more effective and flexible dynamic networks paradigm.

\section{Methodology}
\begin{figure*}[t]
\begin{center}
\includegraphics[width=0.95\textwidth]{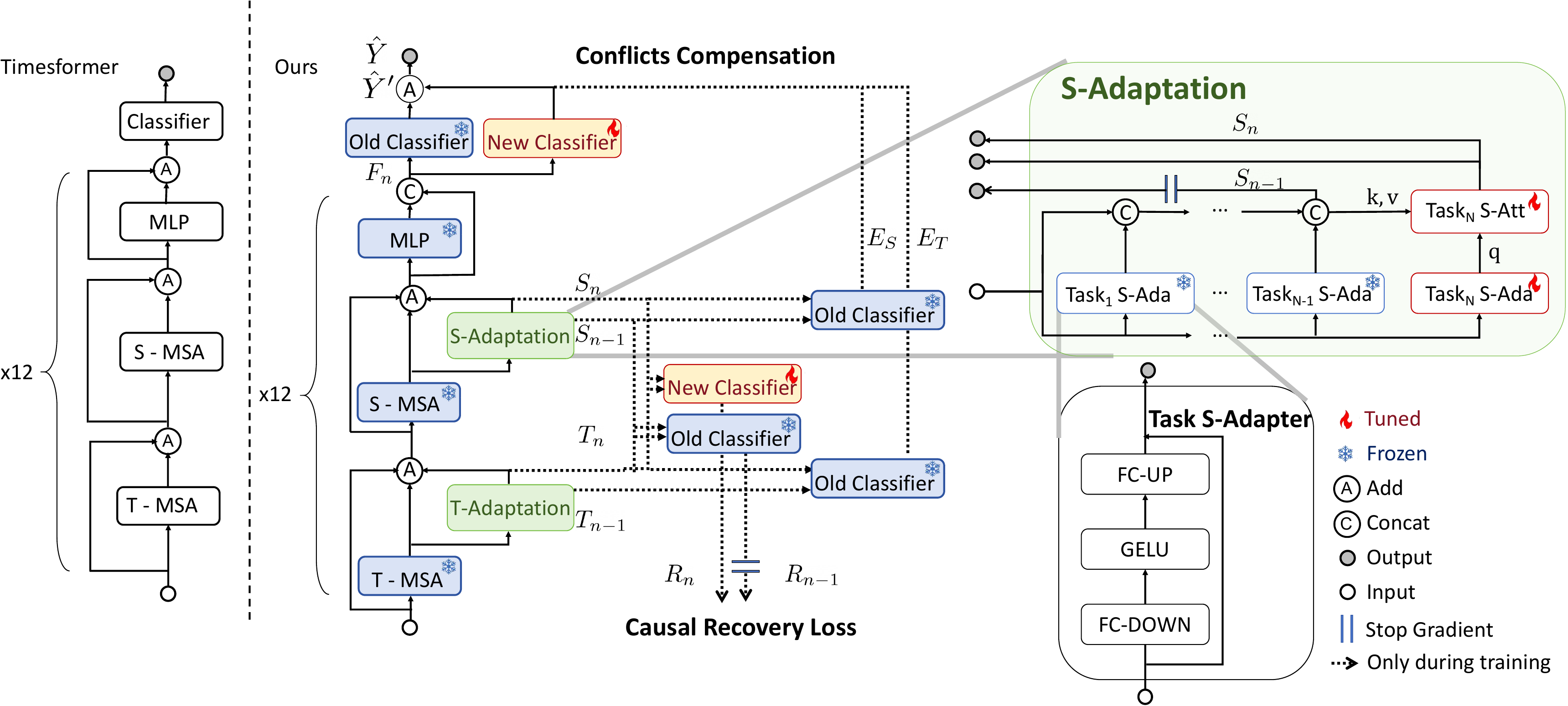}
\end{center}
\caption{\textbf{CSTA Structure.} The left is TimeSformer architecture, while the middle shows the overall structure of our whole spatial-temporal causal adaptation block. The detailed structures of the adaptation module and adapter module are shown on the right, with only the modules marked with a tuned flag being learnable. The causal recovery loss ensures the effective representation of new knowledge through relation recovery between spatial and temporal knowledge, and the conflict compensation effect is incorporated into the classification results from the main branch to enhance memorization. The causal recovery loss and conflict compensation mechanism are employed only during the training phase.}
\label{fig:adapter}
\end{figure*}
\begin{figure*}[t]
\begin{center}
\includegraphics[width=0.9\textwidth]{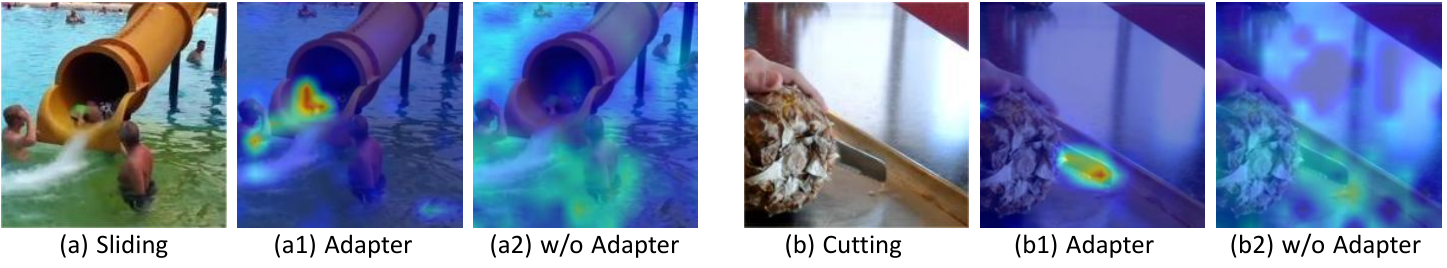}
\end{center}
\caption{\textbf{Visualization.}  Example class (a) is introduced in the new task, (a1) indicates the attention map learned with adapters, while (a2) indicates the attention map extracted by the model trained in old tasks, showing the learning ability of the adapter. In contrast, example class (b) is introduced in the previous task, (b1) indicates the attention map learned with adapters, while (b2) is the attention map extracted by the model with all parameters trainable, showing the memorization ability of the adapter.}
\label{visuall}
\end{figure*}
\subsection{Problem Formulation}
We follow the vCLIMB~\cite{vclimb} benchmark to set up the video CIL task. Specifically, the aim is to train a classification model $f(\cdot)$, learned from $N$ tasks which will arrive in a particular order. 
Each task in the sequence contains its specific dataset $\{task_{0}$ ... $task_{1}$, $task_{N}$\}~=
\{($V_{0}$, $Y_{0}$), ($V_{1}$, $Y_{1}$) ... ($V_{N}$, $Y_{N}$)\} where $V_{i}$ denotes a video sample of class $Y_{i}$. 
The classes of samples introduced at each task are unique, i.e. $Y_{i} \cap Y_{j} = \emptyset, \forall i,j$. 
In our implementation, we adopt the popular transformer-based video network TimeSformer~\cite {timesformer} as $f(\cdot)$.

\subsection{Spatial-Temporal Task Adaptation}
\label{Sec:adapter}

In the realm of image CIL, the dynamic network paradigm has recently attracted significant attention from researchers due to its effectiveness in mitigating forgetting issues without the requirement to store samples. 
However, this approach also carries the risk of introducing a substantial number of parameters as tasks accumulate. Moreover, despite its success in image CIL, this paradigm remains relatively unexplored in the video CIL.
For effective learning and memorizing, in this section, we introduce our proposed lightweight dynamic network based on the TimeSformer backbone. 

Specifically, We argue that the knowledge of video transformers predominantly lies in the patterns of spatial and temporal attention (e.g., the T-MSA and S-MSA in TimeSformer), allowing us to adjust this knowledge through \textbf{\textit{attention expansion}} to suit new tasks while maintaining manageable parameter counts. 
Another significant design consideration is illustrated in (b) and (c) of Fig.~\ref{fig:observasion}, different new action classes exhibit unique preferences for representation increment of spatial and temporal knowledge. 
Consequently, we introduce separate mechanisms for spatial attention expansion and temporal attention expansion.

Inspired by the recent success of adapter-like modules~\cite{pan2022st,yang2023aim} in domain transfer tasks, we introduce a task-specific adapter tailored for the dynamic acquisition of representations from new tasks.
As illustrated in Fig.~\ref{fig:adapter}, when a new task emerges, we incorporate the task-specific adapter module immediately following the MSA (Multi-Head Self-Attention) within each attention block. 
This adapter takes the raw features generated by the raw attention mechanism and learns to modulate these features based on the characteristics of the current task, as shown in Fig.~\ref{visuall}. 
The adapted features are subsequently concatenated with the original features and then passed to the ensuing layers in this block. This adaptation process is conducted over both T-MSA and S-MSA in each block, which can be formulated as:
\begin{equation}
\begin{split}
F_{n}^{MSA} = F_{0}^{MSA} + \sum_{i=1}^{n}
\text{Adapter}^{i}(F_{0}^{MSA})~,\\
\text{Adapter}^{i} = [ W^{i}_2(\text{GeLU}(W^{i}_1(\cdot)) ]
\end{split}
\end{equation}
where $F_{n}^{MSA}$ denotes the output feature of MSA of the last transformer layer ($\text{layer}_{12}$) in $task_{n}$, $W_1, W_2$ denotes weights of linear layer, and $\text{GeLU}$ denotes nonlinear activation function. 
In line with the standard procedure in video CIL, a new linear classifier will also be appended to the existing classification head to recognize classes in the current task. 
Consequently, during the execution of the current task $n$, only the recently introduced adapter, and classifier are open to training, while all other components remain frozen.

To further mitigate the forgetting of previous tasks in the attention representation, we incorporated cross-task attention to facilitate the merging between old tasks and new tasks. 
Specifically, the attention mechanism is performed between all the modulated representations from the previous and the current:
\begin{equation}
\begin{split}
K, V= (F_{0}^{MSA} \oplus \ldots \oplus F_{n-1}^{MSA})~, \sigma =  {\sqrt{{d}/{h}}}~,\\
ACT = \text{{Softmax}}\left({{W_q F_{n}^{MSA} \cdot W_k K^T}}/{\sigma}\right) \cdot W_v V
\end{split}
\end{equation}
where $(F_{0}^{MSA} \oplus \ldots \oplus F_{n-1}^{MSA})$ denotes concatenate all tasks information before along the embedding dimension $D$, $d$ is the embedding dimension, and $h$ is the number of attention heads, and $W_o, W_k, W_v$ are weights of linear project layers. 
Secondly, following the knowledge distillation-based methods in CIL, we conduct a logit-based knowledge distillation for better knowledge memorization. 
Specifically, given the predicted logits of task $n$ classifier under two circumstances: (1) network without adding adapter for adapting to $n$-th task.
(2) network after adapting to $n$-th task, the distillation loss is formulated as:
\begin{equation}
    \mathcal{L}_{D} = loss_{kl}(  \text{Clf}_n(F_{n}), \text{Clf}_n(F_{n-1}) )
\end{equation}
where $\text{Clf}_n$ denotes the classifier in $task_{n}$, $F_{n}, F_{n-1}$ denotes the whole features extracted in $task_{n}, task_{n-1}$. It mitigates forgetting by learning from previous representations.

\subsection{Delving into the Spatial and Temporal Knowledge}
\label{Sec:analyze}

\begin{figure*}[!t]
\begin{center}
\includegraphics[width=0.95\textwidth]{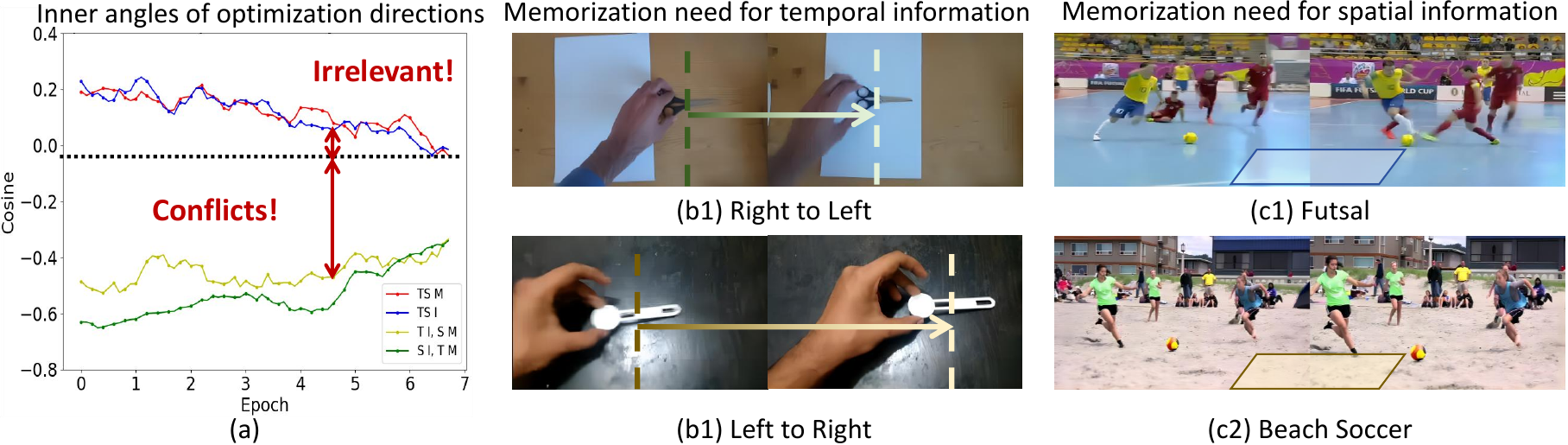}
\end{center}
\caption{\textbf{Motivations.} (a) shows the relation between spatial/temporal increment/memorization in the training process, the \textcolor{blue}{blue line} shows the cosine value between spatial increment and temporal increment in approaching zero which indicates becoming irrelevant. While the cosine value in \textcolor{mycolor_green}{green line} and \textcolor{mycolor_yellow}{yellow line} below zero indicates the conflicts' existence, and the \textcolor{mycolor_red}{red line} indicates the benefit between memorizations can be better used for memory enhancement; (b1) shows the action ``Pulling something from left to right", and (b2) shows the action ``Pulling something from right to left", which exhibit significant different temporal information, in contrast, (c1) shows the action ``Beach soccer", and (c2) shows the action ``Futsal", which exhibit significant different spatial information, so different needs for representation drive our adaptation design to memorize information separately in video CIL.}
\label{fig:observasion}
\end{figure*}

As illustrated in Fig~\ref{fig:observasion}~(b), due to the unique representation demand during an increment, temporal and spatial information are separately enhanced by the aforementioned attention expansion structure. 
However, it is natural to wonder whether the separation of spatiotemporal attention and adapters might also bring some drawbacks, such as the gradual weakening of spatiotemporal relationships, which could make it challenging to maintain a unified and comprehensive representation of action categories. 
To test this hypothesis, we further analyze and quantify the relation between the processes of increment/memorization for spatial and temporal knowledge.

Specifically, we introduce the Kullback-Leibler Divergence between different adaptation implementations. 
When the Kullback-Leibler (KL) loss function is employed to assess the difference between distributions, the optimization objective is to minimize the KL divergence, aligning the target distribution with the source distribution\cite{kl,KLnew}.

Consequently, when the classification results generated by the model with only spatial or temporal adapters are used as the source distribution for the alignment direction, while the model without adapters serves as the target input, the optimization direction of the KL loss indicates an increment process of spatial or temporal information. 
This approach aims to align the distribution of action recognition results from the model without the adaptation module to those with the module, effectively representing an incremental process.

Conversely, when the classification results from the model only with spatial or temporal adapters serve as the source distribution for the alignment direction, while the model with complete spatiotemporal adapters serves as target input, the optimization direction of the KL loss points towards the temporal or spatial memorization process. 
Similar to the analysis of the incremental process, this objective is to align the distribution of action recognition results with the adaptation module to the case without the adaptation module, which is equivalent to a memorization process.

Specifically, we compute the cosine similarity between the gradients of temporal and spatial increment/memorization, allowing us to assess whether these gradients are aligned or diverge in the direction of parameter optimization. 
By examining the sign of the cosine similarity, as suggested by~\cite{angle_of_gradient}, we can determine whether two variables are in conflict or whether they mutually reinforce each other. The cosine value represents the relation between them during optimization (\textbf{cooperation} when the cosine value is above zero, \textbf{conflict} when the cosine value is below zero). 

The results of the quantitative analysis are presented in Fig.~\ref{fig:observasion} (a). These observations can be summarized as:
\begin{itemize}
    \item[(1)] \textit{Separated spatial and temporal adaptation gradually impairs the inherent relation between spatial and temporal knowledge.} (as the \textcolor{blue}{ blue line})
    \item[(2)] \textit{There exists conflicts between the increment of one's information and the memorization of the other's.} (as the \textcolor{mycolor_green}{green line} and \textcolor{mycolor_yellow}{yellow line})
\end{itemize}
Observation-(1) is that, despite the simultaneous optimization of temporal and spatial information representations, the separate attention and adapter structures still hinder further association between them. 
Observation-(2) since the generally contradictory directions of memory and incremental learning; thus, the incremental learning of one type of information (either spatial or temporal) will conflict with the memorization process of the other.
Motivated by these, we introduce two causal mechanisms to our framework to mitigate forgetting from a causal perspective. We will elaborate on them in the following sections.

\subsection{Causal Recovery for Spatial-Temporal Relation}
\label{Sec:relation}

\begin{figure}[t]
\begin{center}
\includegraphics[width=0.48\textwidth]{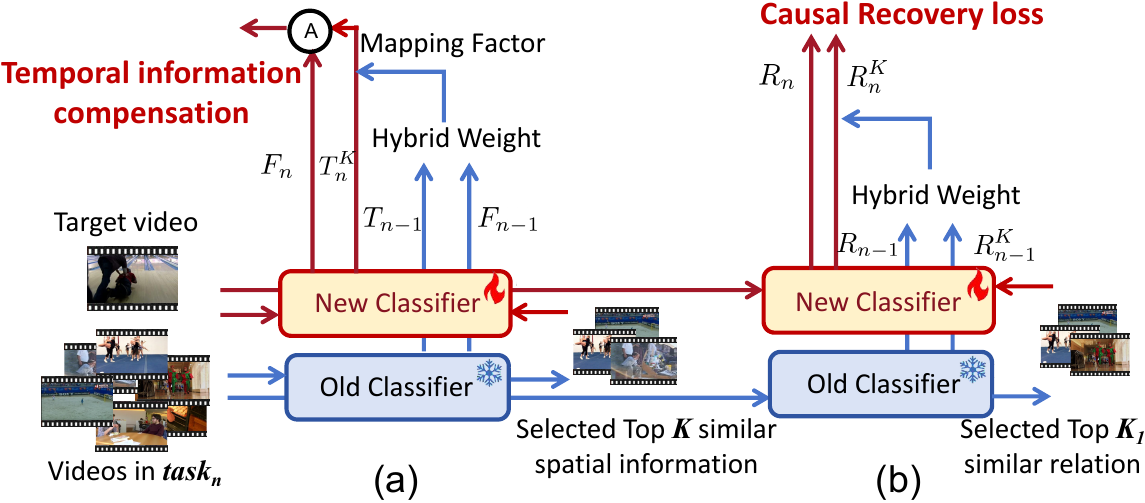}
\end{center}
\caption{\textbf{Causal Methods.} Fig.(a) represents the detailed implementation of relation recovery for the effective representation of new knowledge, while Fig.(b) represents the causal compensation implementation for conflict relieving.}
\label{fig:causal_distillation}
\end{figure}

In class incremental learning, the number of new categories added per task is consistent with $task_0$, but in our attention expansion method, the number of learnable parameters introduced in each task is less than half of those in $task_0$,  underscoring the necessity for a more efficient representation.
As pointed out in~\cite{uniformer}, the efficient representation of information is enhanced when spatiotemporal relationships within a video are more closely intertwined. 
However, as shown in Observation-(1), separated spatial and temporal adaptation gradually impairs the inherent relation between spatial and temporal knowledge.
Therefore, we introduce a causal distillation mechanism to preserve the inherent spatial-temporal relationship.
As shown in Fig.~\ref{fig:causal_distillation} (a), it is achieved by recovering the relation between spatial and temporal knowledge learned in old tasks. 
By doing so, we aim to maintain the similarity between the newly acquired spatial-temporal relation and the original relation, thereby enabling a more efficient representation of the new knowledge while maintaining the original knowledge.

Specifically, we initially represent the relation learned up to the $task_i$ as $R_{i}$, where $i\in [1,2,\dots, N]$. 
We estimate $R_i$ based on the proportion of spatial and temporal information of the classification results of whole spatial-temporal feature $\text{Clf}_{i}(F_{i})$. 
Moreover, as an additional indicator, cosine similarity between the classification results of spatial features $S_i$ and temporal features $T_i$ of the corresponding MSA module in $task_i$ is also concatenated. $R_i$ is represented as spatial relation $R_{S_i}$ and temporal relation $R_{T_i}$, consequently ${R}_{i}= (R_{S_i}, R_{T_i})$ where 
\begin{equation}
\begin{split}
    {R}_{{S_{i}}} = \texttt{{Cat}}\left(\frac{\text{Clf}_{i}(S_{i})}{\text{Clf}_{i}(F_{i})}, \cos\left<\text{Clf}_{i}(S_{i}), \text{Clf}_{i}(T_{i})\right>\right), \\
    {R}_{{T_{i}}} = \texttt{{Cat}}\left(\frac{\text{Clf}_{i}(T_{i})}{\text{Clf}_{i}(F_{i})}, \cos\left<\text{Clf}_{i}(S_{i}), \text{Clf}_{i}(T_{i})\right>\right). 
\end{split}
\label{r_relation}
\end{equation}
Given a certain video sample in $task_n$, other videos in $task_n$ that share similar spatial-temporal relations in previous $task_0~...~task_{n-1}$ serve as a constraint to guide the representation of the relation. The effect of the relation-guided constraint can be represented by conditional probability:
\begin{equation}
\begin{split}
E_{R} = \sum_{v_n} P(\hat{Y_n}|V_n,{R}_{{S_{n}}},{R}_{{T_{n}}}) P({R}_{{S_{n}}},{R}_{{T_{n}}}| R_{{S_{n-1}}}, {R}_{{T_{n-1}}}) 
\label{er}
\end{split}
\end{equation}
where $V_n$ indicates video samples in $task_n$, $\hat{Y_n}$ indicates the classification result. As illustrated in Sec.~\ref{Sec:adapter}, The weight $P({R}_{{S_{n}}},{R}_{{T_{n}}}|R_{{S_{n-1}}}, {R}_{{T_{n-1}}})$ is naively realized through the attention expansion mechanism. 
However, as stated in the observation-(1) in the above section, the temporal and spatial increment is gradually becoming less relevant. A straightforward approach to recovering these relations is to design a recovery loss as follows:
\begin{equation}
\mathcal{L}_T = 1 - \texttt{cos}[{R}_{{T_{n-1}}}, {R}_{{T_n}}],~\mathcal{L}_S = 1 - \texttt{cos}[{R}_{{S_{n-1}}}, {R}_{{S_n}}]
\label{simple}
\end{equation}
where \texttt{cos}(-) refers to the cosine similarity. However, different data distribution may lead this naive distillation loss to mislead the new training~\cite{mislead}. To avoid misleading, we combine the simple loss function in eq.~\ref{simple} with the formula established in eq.~\ref{er} according to causality. 
Inspired by~\cite{hu2021distilling}, The weight $P({R}_{{S_{n}}},{R}_{{T_{n}}}| R_{{S_{n-1}}}, R_{{T_{n-1}}})$ in eq.~\ref{er} can be converted to the measurement of the similarity between samples, where the similarity as a hybrid weight, forming a fusion distillation to ameliorate the negative effects of naive distillation. Therefore the loss function with the hybrid relation consisting of similar samples in the $\text{task}_{(n-1)}$ can be converted to:
\begin{equation}
\mathcal{L}_T = 1 - \texttt{cos}[{R}^{K}_{{T_{n}}}, {R}_{{T_n}}], \mathcal{L}_S = 1 - \texttt{cos}[{R}^{K}_{{S_{n}}}, {R}_{{S_n}}] 
\end{equation}
 where $K$ indicates a hyperparameter of \textit{Top-K} similar relations in the input of current $task_i$, ${R}^{K}_{{T_{n}}}$ and ${R}^{K}_{{S_{n}}}$ indicate the relation mixed by $K$ videos which have a similar relation in $task_{n-1}$. The mixed spatial feature $S^{K}_{n}$ and mixed temporal feature $T^{K}_{n}$ can be calculated by:
\begin{equation}
\begin{split}
S^{K}_{n} = \sum_{k=0}^{K} \text{Clf}_{n}(S^{k}_{n}) \cdot \text{sim}( {R}_{{S^{k}_{n-1}}}, {R}_{{S_{n-1}}}), \\
T^{K}_{n} = \sum_{k=0}^{K} \text{Clf}_{n}(T^{k}_{n}) \cdot \text{sim}( {R}_{{T^{k}_{n-1}}}, {R}_{{T_{n-1}}}).
\end{split}
\end{equation}
where $k$ indicates a certain similar sample. Consequently, the ${R}^{K}_{{T_{n}}}$ and ${R}^{K}_{{S_{n}}}$ can be calculated according to eq.~\ref{r_relation}.

\subsection{Causal Compensation for Spatial-Temporal Conflicts}

In the above section, we strengthen the relation between spatial and temporal information. However, as the Observation-(2), there exist conflicts between spatial increment (or memorization) and temporal memorization (or increment), in contrast, as the \textcolor{mycolor_red}{red line} in (a) of Fig.~\ref{fig:observasion}, there exists a benefit between spatial memorization and temporal memorization. 

Therefore, in addition to enhancing the effect of spatial/temporal own memory by introducing the separate adaptation module, we also introduce the positive effect of the other information's memorization process to compensate for the negative impact caused by its increment as shown in Fig~\ref{fig:causal_distillation} (b). 
Take the temporal information as an example, we construct the causal effect $E_T$ for compensation following~\cite{mecd}, which can be represented by:
\begin{equation}
E_T = \underbrace{E_{S_{n-1}}\vphantom{\frac{\nabla S_{n}}{\nabla T_{n}}}}_{\text{Spatial Benefits}} \hspace{-0.2cm}\cdot \hspace{0.2cm} \underbrace{\frac{\nabla S_{n}}{\nabla T_{n}} \cdot \cos(\nabla S_{n},\nabla T_{n})}_\text{Mapping Influence Factor}
\end{equation}
where $S_{n}$ and $T_{n}$ indicate spatial and temporal features of $task_n$, $\nabla S_{n}$ and $\nabla T_{n}$ indicate the numerical magnitude of the gradient of the spatial increment and temporal increment. $\cos(\nabla S_{n},\nabla T_{n})$ refers to the cosine value of the included angle. For simplicity, the mapping influence factor is simplified as $\alpha_T$.  Similar to eq.~\ref{er}, $E_T$ can be calculated as:
\begin{equation}
{E_T} = \alpha_T  \cdot \sum_{v_n} P(\hat{Y_n}|V_n,T_{n},S_{n})P(T_{n}|S_{n-1})~.
\end{equation}
 Similar to the above section, weights can be converted to hybrid weight which is the measurement of the similarity:
 \begin{equation}
\begin{split}
E_T=\sum_{k=0}^{K_{1}}\text{Clf}_n(T_{n}^k) \cdot \text{sim}(\text{Clf}_{n-1}(F_{n-1})-\text{Clf}_{n-1}(T_{n-1})).
\end{split}
\end{equation}
Similarly, \textit{Top-K} similar temporal features are aggressed to implement $E_T$, where $K_1$ is a hyperparameter. In the same way, the spatial causal compensation effect is represented by: $E_S = \alpha_S \cdot E_{T_{n-1}}$, and is implemented in the same way. 
These effects represent the mixed causal classification results and are compensated to the non-causal classification result $\hat{Y}^{\prime}$ which is illustrated in eq.~\ref{pred} below.
\begin{table*}[t]  
\begin{center}
\caption{\textbf{Main Results.} Our CSTA reports SOTA on $\text{Acc}_N$ for all Activitynet and Kinetics tasks under the balanced splitting manner in the vCLIMB Benchmark, outperforming the previous SOTA by up to \textbf{4.2\%} on average. Besides, CSTA outperforms the previous SOTA by 2.9\% on average at $\bf{\overline{\text{Acc}}}$ for the unbalanced task in the UCF101 and HMDB51 datasets, and reaches the second-best performance on the SSV2 dataset following the setting in the TCD Benchmark.}
\resizebox{0.98\linewidth}{!}{
\setlength{\tabcolsep}{1.2mm}{
\begin{tabular}{p{3.1cm}|ccccccccc|ccc}
\toprule
\multirow{3}{*}{\textbf{Method}} &\multirow{3}{*}{\textbf{Publisher}} & \multicolumn{8}{c|}{\textbf{vCLIMB Benchmark}}  & \multicolumn{3}{c}{\textbf{TCD Benchmark}}\\
{} & & \multicolumn{2}{c}{\textbf{ActivityNet-10}} & \multicolumn{2}{c}{\textbf{ActivityNet-20}} & \multicolumn{2}{c}{\textbf{Kinetics400-10}} & \multicolumn{2}{c|}{\textbf{Kinetics400-20}} & \multicolumn{1}{c}{\textbf{UCF101}} & \multicolumn{1}{c}{\textbf{HMDB}} & \multicolumn{1}{c}{\textbf{SSV2}} \\
{} & & $\textbf{Acc}_N \bf{\uparrow}$ & \textbf{BWF $\downarrow$} & $\textbf{Acc}_N \bf{\uparrow}$ & \textbf{BWF $\downarrow$} & $\textbf{Acc}_N \bf{\uparrow}$ & \textbf{BWF $\downarrow$} & $\textbf{Acc}_N \bf{\uparrow}$ & \textbf{BWF $\downarrow$} & $\overline{\textbf{Acc}} \uparrow$  & $\overline{\textbf{Acc}} \uparrow$  & $\overline{\textbf{Acc}} \uparrow$  \\
\midrule
baseline~(TimeS) &- & 33.22 & 32.92 & 30.68 & 35.80 & 30.31 & 30.46 & 28.44 & 34.51 &63.32 &41.07   & 27.22\\
baseline~(TSN)  &- & 31.97 & 34.61 & 27.09 & 38.06 & 29.68 & 30.97 & 27.03 & 36.14 &63.41 &39.92    & 26.58 \\
TCD~(TSM)~\cite{tcd} & CVPR 2021 & - & - & - & - & - & - & - & - &78.13 &50.36  &35.78 \\
FrameMaker~(TSM)~\cite{framemaker} &NeurIPS 2022& - & - & - & - & - & - & - & - &78.13 &47.54 &37.25 \\
HCE~(CNN)~\cite{hce} &AAAI 2024 & - & - & - & - & - & - & - & - &79.12 &48.63 &38.67  \\
STSP~(TSM)~\cite{stsp} &ECCV 2024 & - & - & - & - & - & - & - & - &81.15 &56.99 &\cellcolor{green!20}\textbf{\underline{69.68}} \\
ST-prompt~(VIT)~\cite{st_prompt} &CVPR 2023 & - & - & - & - & - & - & - & - &83.74 &55.13 &36.84 \\
ST-prompt~(CLIP)~\cite{st_prompt} &CVPR 2023 & - & - & - & - & - & - & - & - &\cellcolor{green!5}\textbf{84.75} &\cellcolor{green!5}\textbf{60.14} &39.98 \\
vCLIMB+iCaRL~(TSN)~\cite{vclimb}  &CVPR 2022 & 48.53 & 19.72 & 43.33 & 26.73 & 32.04 & 38.74 & 26.73 & 42.25 &-&- & - \\
vCLIMB+BiC~(TSN)~\cite{vclimb}  &CVPR 2022 & 51.96 & 24.27 & 46.53 & 23.06 & 27.90 & 51.96 & 23.06 & 58.97 &-&- & - \\
Teacher-Agent~(TSN)~\cite{teacher} &ArXiv 2023&-&-&-&- &38.65 &22.48 &35.38 &30.78 &-&- & -\\
SMILE+iCaRL~(TSN)~\cite{glimpse2023} &CVPR 2023 & 50.26 & 15.87 & 43.45 & 23.02 & 46.58 & 7.34 & 45.77 & \cellcolor{green!5}\textbf{4.57} &- & - & - \\
SMILE+BiC~(TSN)~\cite{glimpse2023} &CVPR 2023 & \cellcolor{green!5}\textbf{54.83} & 7.69 & \cellcolor{green!5}\textbf{51.14} & \cellcolor{green!20}\textbf{\underline{1.33}} & \cellcolor{green!5}\textbf{52.24} & 6.25 & \cellcolor{green!5}\textbf{48.22} & \cellcolor{green!20}\textbf{\underline{0.31}} &- & - & -\\
\midrule
\textbf{CSTA~(Vivit)} &- & 58.31 & 7.06 & 54.96 & 7.51 & 54.98 & 5.06 & 51.01 & 6.91 &87.32 &61.80    & 39.09\\
 \textbf{CSTA~(TimeS)} &- & \cellcolor{green!20}\textbf{\underline{59.24}} & \cellcolor{green!20}\textbf{\underline{6.91}} & \cellcolor{green!20}\textbf{\underline{55.72}} & \cellcolor{green!5}\textbf{7.53} & \cellcolor{green!20}\textbf{\underline{56.09}} & \cellcolor{green!20}\textbf{\underline{4.97}} & \cellcolor{green!20}\textbf{\underline{52.20}} & 6.86  
&\cellcolor{green!20}\textbf{\underline{87.53}}
&\cellcolor{green!20}\textbf{\underline{63.16}}   &\cellcolor{green!5}\textbf{41.26}\\
\bottomrule
\end{tabular}
}}
\label{tab:main}
\end{center}
\end{table*}
\subsection{ Training Objective }
The final output of classification result $\hat{Y}$ and the loss function $\mathcal{L}oss$ utilized while training are:
\begin{gather}
\begin{split}
\label{pred}
\hat{Y} = \hat{Y}^{\prime} + \lambda_1 E_S + \lambda_2 E_T~,\\
\mathcal{L}oss = \mathcal{L}_{CE} + \mu_1 \mathcal{L}_{D} + \mu_2 \mathcal{L}_{T} + \mu_3 \mathcal{L}_{S}~,
\end{split}
\end{gather}
where $\lambda_1$, $\lambda_2$, $\mu_1$, $\mu_2$, and $\mu_3$ are the hyperparameters, for the tradeoff between learning new knowledge and maintaining the old. $\hat{Y}^{\prime}$ is the non-causal classification result, $\mathcal{L}_{CE}$ is classification loss calculated as $\mathcal{L}oss_{CE}(\hat{Y}, Y)$, where $Y$ is the ground truth label and $\mathcal{L}oss_{CE}$ is the cross-entropy loss. 
During the inference phase, it is noteworthy that the relation recovery and conflict compensation mechanisms are not employed, which owing to the relation and conflict have already been recovered and relived in the training phase. Instead, the model generates aggregated adaptation outcomes solely, ensuring a more efficient inference process.

\section{Experiments}
\subsection{Experimental Setup}
\noindent\textbf{Datasets.}
Following the vCLIMB benchmark~\cite{vclimb} and TCD benchmark~\cite{tcd}, we conduct experiments on all action recognition datasets including ActivityNet~\cite{activitynet}, Kinetics400~\cite{kinetics}, HMDB51~\cite{hmdb}, UCF101~\cite{ucf101}, and Something-SomethingV2~\cite{ssv2}, which comprise 200, 400, 51, 101, 174 action classes, respectively.

\noindent\textbf{Task Split.}
There are two task-splitting manners for video CIL: 1. balance splitting all classes. 2. splitting half of the classes first and then balance splitting the remaining classes. To better compare with previous SOTA methods, we follow the splitting method corresponding to the dataset. Different splitting manners can also better evaluate generalizability.

For the balanced splitting manner, to align with the vCLIMB, for ActivityNet and Kinetics400 datasets, all classes are evenly divided into 10 or 20 tasks. Additionally, for the unbalanced splitting manner, to align with the TCD, for the UCF101 and SSV2 datasets, we learn 51 and 84 classes in the first task, the remaining 50 and 90 classes are evenly divided into 10 tasks. Similarly, for the HMDB51 dataset, aligning with TCD, we learn 26 classes first, and the remaining 25 classes are evenly divided into 5 tasks. 

\noindent\textbf{Evaluation Metrics.}
We utilize the Top-1 accuracy after final $task_N$ as $\text{Acc}_N$ to evaluate on ActivityNet and Kinetics400, aligning with Teacher Agent, vCLIMB, and SMILE. 
Additionally, aligning with vCLIMB, we employ Backward Forgetting (BWF) to evaluate the influence of learning new tasks, $\text{BWF}~=\frac{1}{N-1}\sum_{i=1}^{N-1}(\text{Acc}_{i}-\text{Acc}_{N})$ where $\text{Acc}_{i}$ indicates the accuracy after $task_i$. 
While the average Top-1 accuracy $\overline{\text{Acc}}$ is utilized for the UCF101, SSV2, and HMDB51 datasets, aligning with TCD, FrameMaker, and ST-Prompt, $\overline{\text{Acc}}$ is calculated by the equation: $\overline{\text{Acc}}~=\frac{1}{N-1}\sum_{i=1}^{N-1}\text{Acc}_{i}$.

\noindent\textbf{Implementation Details.}
We conduct experiments on our proposed CSTA with Vivit~\cite{arnab2021vivit} and Timesformer~\cite{timesformer} as the backbone. Besides, we conduct a fine-tuning process after each training phase. During fine-tuning, we randomly select 5 examples from each class in previous tasks to construct a balanced fine-tuning set, aligning with the definition in the TCD Benchmark. 
During the fine-tuning period, the classifier is tuned while the feature extractor remains frozen.
The fine-tuning method is widely utilized and accepted in knowledge distillation-based and dynamic network-based works in class incremental learning, such as TCD~\cite{tcd}, FA~\cite{fa}, CCIL~\cite{CCIL}, DER~\cite{yan2021dynamically}, Dytox~\cite{douillard2022dytox} and DNE~\cite{hu2023dense}. 
We conducted our experiments on 4 NVIDIA A6000 GPUs, with 5 epochs for training and 1 epoch for fine-tuning in each task. The longest training duration is 12 hours for the Kinetics400-20 tasks, whereas the shortest is 40 minutes for the UCF101 dataset. The remaining datasets or settings span training durations between these two extremes.

\noindent\textbf{Hyperparameters.}
For Kinetics and HMDB51, $\lambda_1$ and $\lambda_2$ are set to 0.1, while for SSV2, UCF101, and ActivityNet, they are set to 0.2. 
Similarly, $\mu_1$, $\mu_2$ and $\mu_3$ are set to 0.1 for Kinetics and HMDB51, and 0.15 for SSV2, UCF101, and ActivityNet, gradually decreasing during training.
The hyperparameters $K$ and $K_1$ are set to 5 for the trade-off of training time-consuming and accuracy.

\subsection{ Comparison to the State of the Art}
\label{sec: main_results}
As shown in Tab.~\ref{tab:main}, We compare our work with all existing works. To indicate the transformer baseline's fairness, we adopt TSN(resnet34)~\cite{wang2018temporal} as the backbone for comparison. The comparable results indicate the fairness of our transformer backbone.
Additionally, to indicate the robustness of our method, we evaluate our CSTA based on Vivit~\cite{arnab2021vivit}, reaching a comparable result to the TimeSformer base. 
In the baseline, we set the MLP of every block trainable and employed the distillation method with KL-loss following the baseline setting in dynamic network-based works~\cite{foster}. 
Overall, our CSTA has a significant improvement over the baseline setting and performs SOTAs on all datasets with both balanced and unbalanced splitting manners. 

\noindent\textbf{10-tasks vCLIMB Benchmark.} 
For the ActivityNet dataset, the results show that CSTA improves over the previous SOTA~ Teacher-Agent~\cite{teacher} and Glimpse~\cite{glimpse2023}. Besides, CSTA also shows an improvement in the Kinetics400 dataset, improving by 3.85\% over the previous SOTA. Additionally, CSTA exhibited SOTA performance on the BWF metric, indicating the strong ability to resist forgetting.

\noindent\textbf{20-tasks vCLIMB Benchmark.} Additionally, we conclude the SOTA comparison by addressing the more challenging 20-tasks setting. Despite the increased difficulty, we observe that CSTA still performs the SOTA on all datasets by up to 4.58\% (ActivityNet). The Backward Forgetting (BWF) is also close to the SOTA at all 20-task settings, the BWF metrics do not perform as well as the 10-tasks may be due to the fact that compared to video data replay, although our designed attention expansion is efficient and memorable, this method is still indirect. However, we still perform SOTA at accuracy without privacy issues.

\noindent\textbf{TCD Benchmark.} 
Additionally, we also reach the second-best performance on the SSV2 dataset, even surpassing the ST-prompt method with a more robust CLIP base. 
Furthermore, we attain the SOTA results on the UCF101 and HMDB51 datasets, the evaluation on the TCD Bench suggesting that our approach is also effective with unbalanced splitting manners.

\subsection{Ablation Studies}

During the ablation study, we firstly validate the effectiveness of our key components in Tab.~\ref{tab:key}, for further proof, as shown in Tab.~\ref{tab:causalana}, we conduct an extra challenging test. 
\begin{table}[t]
\centering
\caption{\textbf{Ablation Study.} We evaluate our causal methods respectively on ActivityNet-10 tasks, where Sep-Ada indicates spatiotemporal separate adaptation, RR indicates the relation recovery, CC indicates the conflict compensation.}
\resizebox{0.95\linewidth}{!}{
\setlength{\tabcolsep}{2.0mm}{
\begin{tabular}{ccc|ccccc}
\toprule
\multicolumn{3}{c}{\textbf{Method}} & \multicolumn{5}{c}{\textbf{Task Order of ActivityNet-10 tasks}}\\
Sep-Ada &RR &CC &1 &3  &5 &7 &9\\
\midrule
& &   & 86.59 & 53.32 & 41.91 & 35.52 & 33.22 \\
\checkmark & &   & 88.44 & 58.39 & 46.40 & 43.86 & 40.46 \\
\checkmark &\checkmark &   & 89.28 & 56.04 & 50.25 & 47.39 & 44.61 \\
\checkmark & & \checkmark   & 91.63 & 60.21 & 54.84 & 47.98 & 44.51 \\
\checkmark &\checkmark & \checkmark  & 92.87 & 72.37 & 62.38 & 58.89 & 59.04 \\
\bottomrule
\end{tabular}}}
\label{tab:key}
\end{table}

\begin{figure}[!tbp]
\begin{center}
\includegraphics[width=0.4\textwidth]{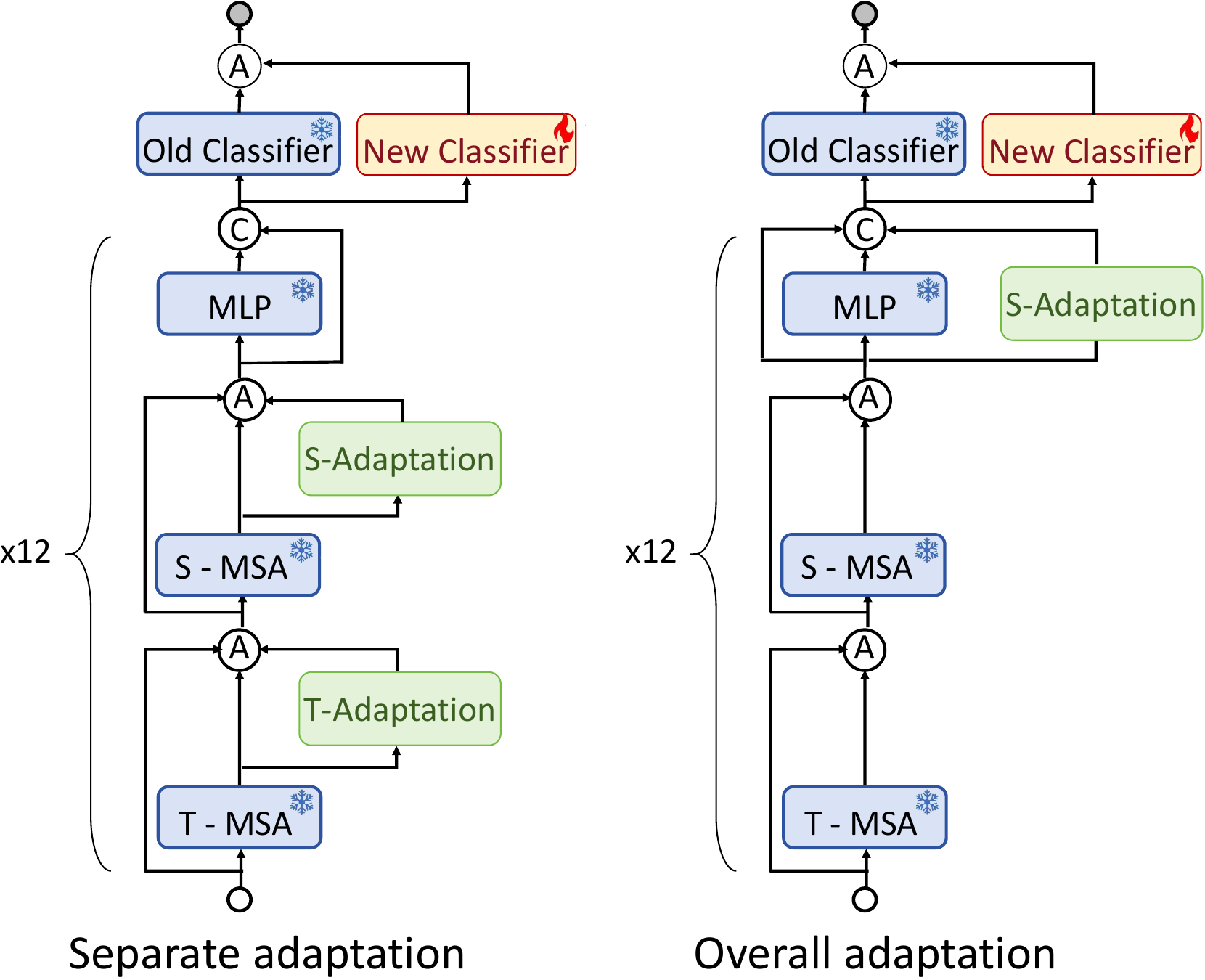}
\end{center}
\caption{\textbf{Adaptation Method Comparison.} We compared the different performances of our current spatiotemporal separate adaptation on the left with the overall adaptation method on the right.}
\label{fig: separate}
\end{figure}

\noindent\textbf{Spatiotemporal Separate Adaptation.} We evaluate the contribution of our separate adapter design in addressing the distinct requirements for different classes during the increment. 
As shown in Fig~\ref{fig: separate}, we compare the performance of a separate adapter setup with that of employing a single adapter for the MLP layer as suggested in \cite{yang2023aim}, which treats video information as a unified whole. 
The observed improvement in performance in Tab.~\ref{tab:key} indicates that the introduction of lightweight separate adapter modules is effective in satisfying the unique demands for spatiotemporal information during the increment.

\noindent\textbf{Relation Recovery.} To demonstrate that the relation recovery can recover the relation learned before, we apply the causal relation recovering method on the base adapter module, the performance increases by 4.15\% in Tab.~\ref{tab:key}, showing that intrinsic relation recovering is necessary for the case of separate adapter modules. Additionally, when we replace the relation recovery loss with traditional loss utilized in knowledge distillation-base and dynamic network-based methods in CIL, such as KL-loss or simple cosine similarity loss to supervise the relation, the accuracy also drops by 1.96\% and 1.77\%.

\noindent\textbf{Conflict Compensation.} To show that causal compensation can relieve conflicts, we adopt spatial and temporal information compensation. This resulted in an accuracy increase of 4.05\% in Tab.~\ref{tab:key}, showing that conflicts between spatial and temporal information are suppressed.

\begin{table}[t]
\centering
\caption{\textbf{Further Analysis of Causal Methods.} We demonstrate the effect of causal methods through more challenging test experiments, which exhibit a substantial disparity in video content across different tasks.}
\resizebox{0.95\linewidth}{!}{
\setlength{\tabcolsep}{1.5mm}{
\begin{tabular}{ccc|ccc}
\toprule
Adapter &RR &CC & Ball Game & Food Making & Water Sports \\
\midrule
\checkmark& & &19.05  &29.86   & 4.39\\
\checkmark&\checkmark & &41.48  &47.69   & 12.05\\
\checkmark& &\checkmark &39.71  &45.84   &11.96 \\
\checkmark&\checkmark &\checkmark  & \textbf{\underline{44.62}} & 
\textbf{\underline{49.90}} & \textbf{\underline{13.07}} \\
\bottomrule
\end{tabular}}}
\label{tab:causalana}
\end{table}

\begin{figure}[!tbp]
\begin{center}
\includegraphics[width=0.48\textwidth]{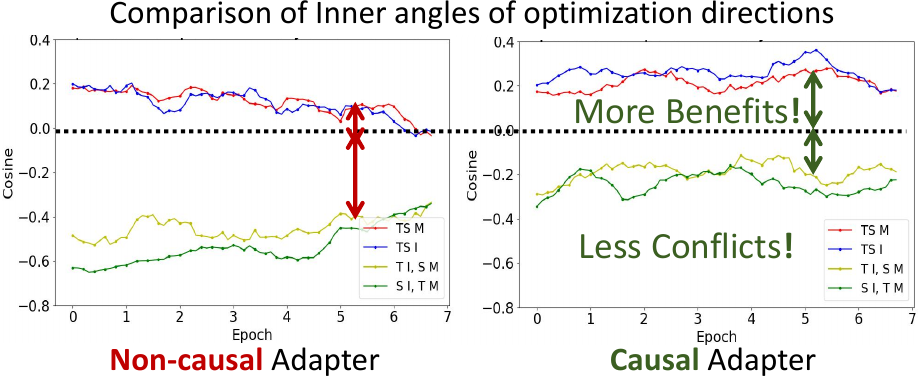}
\end{center}
\caption{\textbf{Effect of Causal Methods.} Visualization of spatiotemporal relations in the training process, the left figure with non-causal adapters while the right one with causal adapters. The \textcolor{mycolor_red}{red line} represents the memorization of both, the \textcolor{blue}{blue line} represents the increment of both, and the else two represent the increment of one and the memorization of the other. With causal recovery, the relation between two increments is closer, and the conflicts are released.}
\label{cauada}
\end{figure}

\subsection{Challenging Setting Test}
The lightweight adapter structure we proposed has already proven effective on two general benchmarks, TCD Bench and vCLIMB. 
Still, we consider whether the method is sufficiently robust in more challenging settings, particularly when there is a significant disparity in the video content learned by the preceding and following tasks.

As shown in Fig.~\ref{tab:causalana}, we split part of the ActivityNet dataset into 3 similar action groups. Each group comprises 8 classes, and we conduct the following protocol for further ablation: 
training 4 classes in $\text{group}_i$ at $task_0$, then training with the other action $\text{group}_j (i \ne j)$ at $task_1$. 
And finally reporting the performance on the remaining 4 classes of the initial $\text{group}_i$ without fine-tuning. 
The results in Tab.~\ref{tab:causalana} indicate that causal methods effectively recover the relation and release the conflicts. 

\begin{figure*}[t]
  \centering
    \begin{minipage}[b]{0.3\textwidth}
    \centering
    \includegraphics[width=1.0\textwidth, height=1.35in]{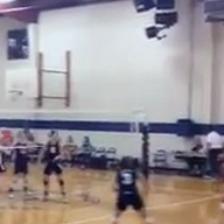}
    \captionsetup{skip=0pt, font=small} 
    \caption*{(a) Volleyball}
  \end{minipage}
  \begin{minipage}[b]{0.3\textwidth}
    \centering
    \includegraphics[width=1.0\textwidth, height=1.35in]{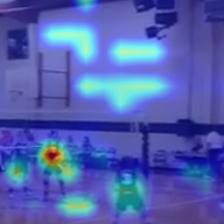}
    \captionsetup{skip=0pt, font=small}
    \caption*{Adapter}
  \end{minipage}
  \begin{minipage}[b]{0.3\textwidth}
    \centering
    \includegraphics[width=1.0\textwidth, height=1.35in]{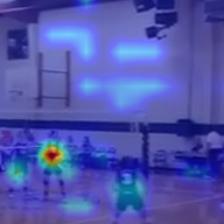}
    \captionsetup{skip=0pt, font=small} 
    \caption*{Causal Adapter}
  \end{minipage}
  \begin{minipage}[b]{0.3\textwidth}
    \centering
    \includegraphics[width=1.0\textwidth, height=1.35in]{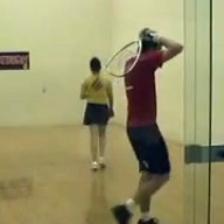}
    \captionsetup{skip=0pt, font=small} 
    \caption*{Playing racquetball}
  \end{minipage}
  \begin{minipage}[b]{0.3\textwidth}
    \centering
    \includegraphics[width=1.0\textwidth, height=1.35in]{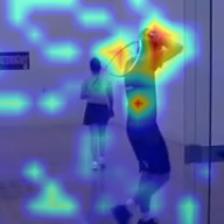}
    \captionsetup{skip=0pt, font=small} 
    \caption*{Adapter}
  \end{minipage}
    \begin{minipage}[b]{0.3\textwidth}
    \centering
    \includegraphics[width=1.0\textwidth, height=1.35in]{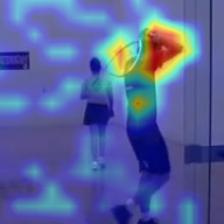}
    \captionsetup{skip=0pt, font=small} 
    \caption*{Causal Adapter}
  \end{minipage}
    \begin{minipage}[b]{0.3\textwidth}
    \centering
    \includegraphics[width=1.0\textwidth, height=1.35in]{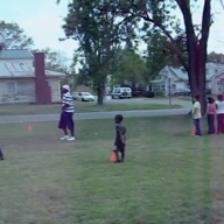}
    \captionsetup{skip=0pt, font=small} 
    \caption*{Playing kickball}
  \end{minipage}
  \begin{minipage}[b]{0.3\textwidth}
    \centering
    \includegraphics[width=1.0\textwidth, height=1.35in]{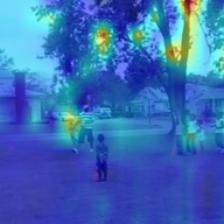}
    \captionsetup{skip=0pt, font=small} 
    \caption*{Adapter}
  \end{minipage}
    \begin{minipage}[b]{0.3\textwidth}
    \centering
    \includegraphics[width=1.0\textwidth, height=1.35in]{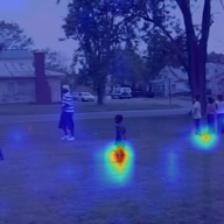}
    \captionsetup{skip=0pt, font=small} 
    \caption*{Causal Adapter}
  \end{minipage}
      \begin{minipage}[b]{0.3\textwidth}
    \centering
    \includegraphics[width=1.0\textwidth, height=1.35in]{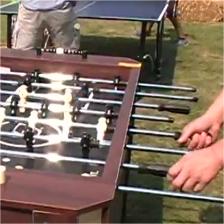}
    \captionsetup{skip=0pt, font=small} 
    \caption*{(d)Table Soccer}
  \end{minipage}
  \begin{minipage}[b]{0.3\textwidth}
    \centering
    \includegraphics[width=1.0\textwidth, height=1.35in]{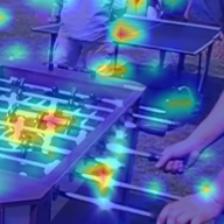}
    \captionsetup{skip=0pt, font=small} 
    \caption*{Adapter}
  \end{minipage}
    \begin{minipage}[b]{0.3\textwidth}
    \centering
    \includegraphics[width=1.0\textwidth, height=1.35in]{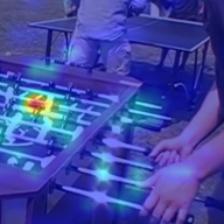}
    \captionsetup{skip=0pt, font=small} 
    \caption*{Causal Adapter}
  \end{minipage}
\caption{\textbf{Effect of Causal Inference.} 
 We visualize a heatmap comparison between the use of causal inference methods and their absence during the challenging test. The adapter-based dynamic model yields cleaner attention maps with causal inference methods, which successfully focus on the subject character and the corresponding action.}
\label{fig:vis_sup}
\end{figure*}

In the challenging setting test, applying separate adapters with causal inference methods yields increases of 25.57\%, 20.04\%, and 8.68\% in TOP-5 accuracy compared to applying the non-causal adapters. As shown in Fig.~\ref{cauada}, when introducing causal methods, the visualization of the cosine value curve indicates that causal methods do relieve conflicts and recover relations.
Additionally, comparisons of heatmap extracted by causal adapters and non-causal adapters are shown in Fig.~\ref{fig:vis_sup}. The heatmap clearly illustrates that the attention module with causal adapters produces a more concentrated attention map, suggesting that causal inference significantly aids in the more efficient memorization of acquired knowledge.

\begin{table}[t]
\centering
\caption{\textbf{Parameters Size Analysis.} Our CSTA model only introduces lightweight adapters, the trainable parameters of the subsequent tasks constitute only 47\% of the initial $task_0$.}
\resizebox{0.95\linewidth}{!}{
\setlength{\tabcolsep}{2.0mm}{
\begin{tabular}{lcccc}
\toprule
Parameters & $task_0$ & $task_1$ & ... & $task_{19}$\\
\midrule
Trainable & 122.47M & 57.34M &   ...   & \textbf{\underline{57.34M}} \\
Total & 122.47M & 236.53M &   ...   & \textbf{\underline{247.46M}} \\
\bottomrule
\end{tabular}}}
\label{tab:memory}
\end{table}

\begin{table}[t]
\centering
\caption{\textbf{Storage Budget Comparison.} We achieved an average storage budget saving of 61.9\% for the Kinetics and ActivityNet datasets compared to the most budget-saving methods previously reported.}
\resizebox{0.95\linewidth}{!}{
\setlength{\tabcolsep}{4.5mm}{
\begin{tabular}{lcc}
\toprule
Method & Kinetics400 & ActivityNet \\
\midrule
SMILE~\cite{glimpse2023} & 3694 MB & 231 MB\\
\textbf{CSTA(FT)} & \textbf{\underline{149(179) MB}} & \textbf{\underline{149(164) MB}}  \\
\bottomrule
\end{tabular}}}
\label{tab:budget}
\end{table}

\begin{table}[t]
\centering
\caption{\textbf{Analysis of Fairness.} The similar results of the previous SOTA SMILE with TSN and Timesformer as backbone indicate the fairness of our transformer-based work.}
\resizebox{0.95\linewidth}{!}{
\setlength{\tabcolsep}{3.0mm}{
\begin{tabular}{lccc}
\toprule
Group & ActivityNet-10 & ActivityNet-20\\
\midrule
SMILE~\cite{glimpse2023} (TSN) &50.26  &43.45 \\
SMILE~\cite{glimpse2023} (TimeS) &51.95 &45.04 \\
\bottomrule
\end{tabular}}}
\label{tab:fairness_ana}
\end{table}

\subsection{Efficiency Analysis}
\label{size}

\noindent\textbf{Parameters.}
As shown in Table \ref{tab:memory}, in addition to the adapter, the introduction of cross-task attention also increases the model's parameters, resulting in a doubling of the parameters in $task_{19}$ relative to $task_0$. However, the trainable parameters constitute only 47\% of those in $task_0$, thus minimizing the computational resources required. 

\noindent\textbf{Storage Budget.}
As illustrated in Table~\ref{tab:budget}, the storage budget for our CSTA encompasses the weights of adapters and cross-task attention, amounting to approximately 99MB and 199MB for 10-task and 20-task settings, respectively, in the vCLIMB benchmark (resulting in an average of 149MB). 
Moreover, the spatiotemporal features of the video do not require separate storage; the cache of the features is purged upon the completion of each task's training.

Currently, SMILE stands as the most storage-efficient, requiring approximately 3694MB for storage memory consisting of replayed video frames. In comparison, our method significantly reduces the average storage to 4.0\% and 64.5\% for the Kinetics400 and ActivityNet datasets, respectively. Even with the inclusion of the fine-tuning process, our method's storage budget of 179MB remains modest, occupying only 5.2\% and 71.0\% of SMILE's storage budget.

\subsection{Fairness Analysis}

As discussed above in Sec.~\ref{sec: main_results}, we conduct our simple fine-tuned baseline on a widely utilized baseline TSN(resnet34) and reach a similar result to the simple fine-tuned TimeSformer baseline. 
To further indicate the fairness of utilizing a transformer baseline, we conduct the previous SOTA SMILE~\cite{glimpse2023} on the TimeSformer base. The results shown in Tab.~\ref{tab:fairness_ana} further indicate the fairness of our transformer-based CSTA over the CNN-based TSN backbone.

\section{Conclusion}
In this work, We reveal the relations between temporal and spatial
information of video during increment tasks. Besides, we introduce CSTA, a simple yet effective causal spatial-temporal adapter-based model. We show that our model outperforms 4.2\% on average over the previous SOTA in the 10 and 20 task settings proposed in the vCLIMB benchmark while only occupying 5.2\% and 71.0\% storage on the Kinetics and ActivityNet datasets, respectively. Additionally, CSTA also achieves SOTA or second-best performances with the challenging unbalanced splitting manner of UCF101, HMDB51, and SSV2 datasets, exhibiting robust generalization capabilities over video class incremental tasks.



\bibliographystyle{IEEEtran}
\bibliography{main}


\vfill

\end{document}